\documentclass[11pt]{article}
\pdfoutput=1
\usepackage{fullpage}
\usepackage{natbib}
\usepackage{hyperref}
\usepackage{url}
\usepackage{enumitem}
\usepackage{tikz}
\usepackage{multirow}
\usepackage{adjustbox}
\usepackage{booktabs}
\usepackage{colortbl}
\usepackage{xcolor}
\usepackage{arydshln}
\usepackage{xspace}
\usepackage{wrapfig}
\usepackage[ruled, linesnumbered, noend, vlined]{algorithm2e}
\usepackage{soul}
\usepackage{tikz}
\usepackage{amsmath}
\usepackage[capitalise]{cleveref}
\usepackage{stackengine}
\usepackage{amsthm}

\usepackage{amsmath,amsfonts,bm}

\def\eqref#1{equation~\ref{#1}}

\def\1{\bm{1}}

\DeclareMathAlphabet{\mathsfit}{\encodingdefault}{\sfdefault}{m}{sl}
\SetMathAlphabet{\mathsfit}{bold}{\encodingdefault}{\sfdefault}{bx}{n}

\def\sL{{\mathbb{L}}}

\def\sR{{\mathbb{R}}}

\newcommand{\data}{\mathcal{D}}
\newcommand{\sw}{w}
\newcommand{\pl}{\mathbf{y}}
\newcommand{\train}{{\mathrm{train}}}
\newcommand{\aug}{{\mathrm{aug}}}
\newcommand{\val}{{\mathrm{val}}}

\newcommand{\batch}{\mathrm{batch}}
\newcommand{\para}{\theta}
\newcommand{\loss}{\mathcal{L}}
\newcommand{\ce}{\text{CE}}
\newcommand{\contrast}{\text{contrast}}
\newcommand{\clip}{\text{CLIP}}
\newcommand{\nn}[1]{f\left(#1\right)}
\newcommand{\cenn}[1]{\bar{\phi}\left(#1\right)}
\newcommand{\ienn}[1]{\bar{g}\left(#1\right)}
\newcommand{\tenn}[1]{\bar{h}\left(#1\right)}
\newcommand{\gip}{\mathbf{\Pi}}
\newcommand{\grvv}{\mathbf{g}}

\newtheoremstyle{break}
  {}
  {}
  {\itshape}
  {}
  {\bfseries}
  {}
  {\newline}
  {}

\newtheorem{theorem}{Theorem}

\newtheorem{notation}{Notation}
\newenvironment{prevproof}[1]{\noindent {\em {Proof of \cref{#1}:}}}{\hfill $\square$\vskip \belowdisplayskip}

\newcommand{\SAFLEX}{\textsc{SAflex}\xspace}
\newcommand{\cm}[1]{\noindent\textbf{#1}}
\definecolor{lightgray}{rgb}{0.9,0.9,0.9}

\setstackgap{L}{1.0\normalbaselineskip}
\stackMath
\usepackage{subcaption}

\title{SAFLEX: Self-Adaptive Augmentation via Feature Label Extrapolation}

\author{
Mucong Ding \and Bang An \and Yuancheng Xu \and Anirudh Satheesh \and Furong Huang\\[2ex]
\normalsize{Department of Computer Science, University of Maryland, College Park, MD, USA}\\[2ex]
\normalsize{\texttt{\{mcding,bangan,ycxu,anirudhs,furongh\}@umd.edu}}
}

\date{}
\begin{document}

\maketitle

\vspace{-1em}
\begin{abstract}
Data augmentation, a cornerstone technique in deep learning, is crucial in enhancing model performance, especially with scarce labeled data.
While traditional methods, such as hand-crafted augmentations, are effective but limited in scope, modern, adaptable techniques often come at the cost of computational complexity and are hard to fit into existing processes.
In this work, we unveil an efficient approach that universally enhances existing data augmentation techniques by enabling their adaptation and refinement, thereby providing a significant and comprehensive improvement across all existing methods.
We present \textbf{\SAFLEX} (\textbf{S}elf-Adaptive \textbf{A}ugmentation via \textbf{F}eature \textbf{L}abel \textbf{EX}trapolation), an approach that utilizes an efficient bilevel optimization to learn the \textit{sample weights} and \textit{soft labels} of augmented samples. This is applicable to augmentations from any source, seamlessly integrating with existing upstream augmentation pipelines.
Remarkably, \SAFLEX effectively reduces the noise and label errors of the upstream augmentation pipeline with a marginal computational cost.
As a versatile module, \SAFLEX excels across diverse datasets, including natural, medical images, and tabular data, showcasing its prowess in few-shot learning and out-of-distribution generalization.
\SAFLEX seamlessly integrates with common augmentation strategies like RandAug and CutMix, as well as augmentations from large pre-trained generative models like stable diffusion. It is also compatible with contrastive learning frameworks, such as fine-tuning CLIP.
Our findings highlight the potential to adapt existing augmentation pipelines for new data types and tasks, signaling a move towards more adaptable and resilient training frameworks.
\end{abstract}
\vspace{-1em}

\vspace{-0.4em}
\section{Introduction}
\label{sec:introduction}
\vspace{-0.4em}

Data augmentation is a cornerstone in improving machine learning models, especially when labeled data is scarce. It enhances model performance by introducing varied training samples. Though traditional methods like rotation and cropping are widely used, they operate under a one-size-fits-all assumption that often falls short in the complexity of real-world data. The key is not just to augment data, but to do it in a way that does not mislead the learning process.



Recent work emphasizes the benefits of learned data augmentation, where techniques such as AutoAugment~\citep{cubuk2019autoaugment} and RandAugment~\citep{cubuk2020randaugment} adapt to specific datasets and tasks. 
While promising, this area is still nascent and lacks a comprehensive framework to address diverse tasks and data nuances. 
Furthermore, selecting meaningful transformations remains a challenge, often relying on heuristics or domain expertise, which is especially problematic in specialized fields. 
Inappropriate transformations can harm model performance, underscoring the need for systematic selection. 
Amid the rise of image generation methods, such as diffusion models and other generative AI, an abundance of synthetic data is available but requires discerning use. 
A recent study, LP-A3~\citep{yang2022adversarial}, aims to generate ``hard positive examples'' for augmentation but risks introducing false positives that could mislead learning.
Another recent work, Soft-Augmentation~\citep{liu2023soft}, introduces soft learning targets and loss reweighting to train on augmented samples but is primarily limited to improving image crop augmentation.
The overarching need is for smarter, more adaptable data augmentation algorithms.

\begin{figure}[!tbp]
    \centering
    \includegraphics[width=0.65\textwidth]{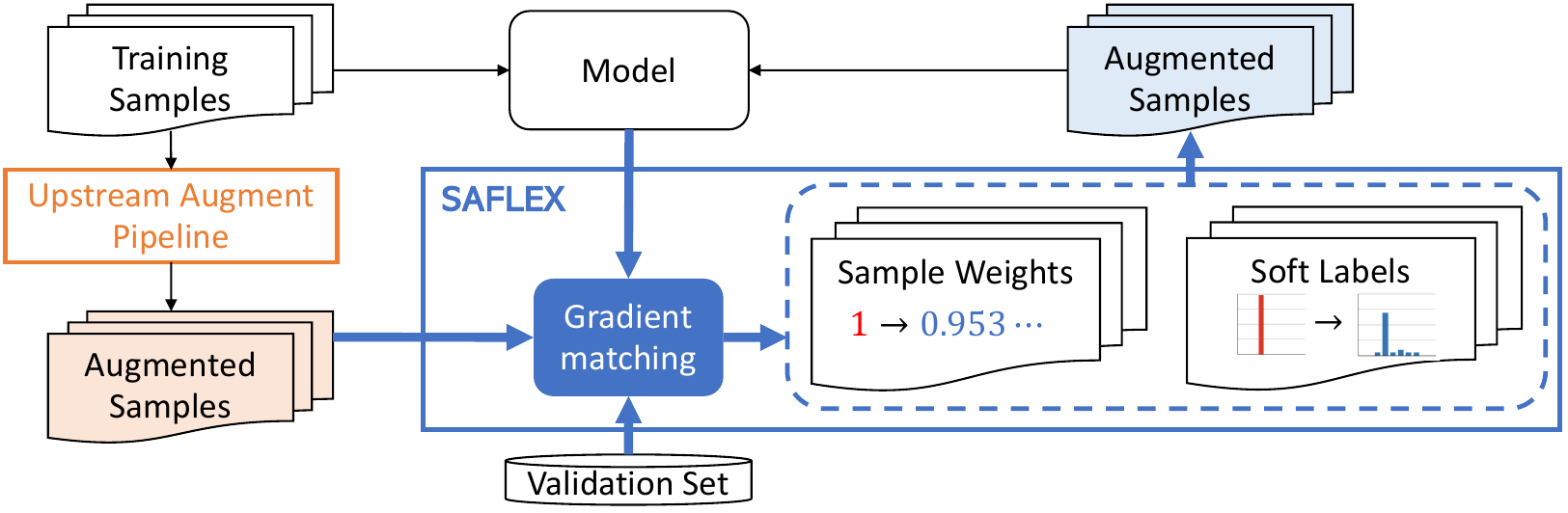}
    \caption{\SAFLEX learns to adjust sample weights and soft labels of augmented samples from an upstream pipeline, aiming to maximize the model's performance on the validation set. While formulated as a bilevel optimization problem, it can be efficiently solved by linear programming with a gradient-matching objective. \SAFLEX is a plug-in to the existing training framework.}
    \label{fig:diagram}
    \vspace{-1em}
\end{figure}

This paper proposes \SAFLEX (Self-Adaptive Augmentation via Feature Label Extrapolation), which automatically learns the sample weights and soft labels of augmented samples provided by any given upstream augmentation pipeline. 
Existing learnable augmentation methods that directly learn in the feature space (e.g., image space) often restrict augmentation scope due to differentiability needs and suffer from complicated training in high-dimensional spaces. 
Contrary to this, we advocate for learning only low-dimensional sample weights and soft labels for each augmented instance sourced from a pre-existing upstream augmentation pipeline like synthetic data generation. 
While upstream augmentation methods can sometimes alter labels or introduce noise, especially when creating samples outside the data distribution, our approach offers a mechanism to correct them.
By calibrating sample weights and labels after augmentation, we considerably alleviate issues stemming from upstream augmentation methods. 
Without the complexity of learning augmentation transformations from scratch, this strategy ensures that augmentation is both diverse and consistent with the inherent data distribution, thereby fostering better generalization across various tasks. See \cref{fig:diagram} for a demonstration of our proposed \SAFLEX.

We frame learning sample weights and soft labels as a bilevel optimization problem. This captures the interdependent nature of the model and its augmented data: the model's performance depends on the quality of the augmented data, which in turn is guided by the model itself~\citep{bard2013practical}. This new perspective advances our understanding of data augmentation, offering a theoretical framework that underpins its practical applications.
Despite the bilevel nature of the problem, direct solutions are computationally infeasible for large-scale real-world applications. To mitigate this, we propose a streamlined, greedy, online, single-level approximation algorithm, which optimizes a gradient-matching objective to accelerate the learning process.

We conducted extensive empirical evaluations to highlight \SAFLEX's superior performance and adaptability.  
On eight medical images~\citep{yang2023medmnist}, \SAFLEX elevates popular augmentation techniques like RandAugment~\citep{cubuk2020randaugment} and Mixup~\citep{zhang2018mixup}, boosting performance by up to $3.6\%$.
On seven tabular datasets, \SAFLEX shows compatibility with categorical data and effectively enhances CutMix~\citep{yun2019cutmix}. 
Furthermore, \SAFLEX improves image augmentations from diffusion models, yielding an average improvement of $1.9\%$ in fine-grained classification and out-of-distribution generalization against three diffusion-augmentation methods, harnessing on their pre-trained expertise.
We also validate \SAFLEX's integration with contrastive learning through a CLIP fine-tuning experiment.
These findings underline its versatility across varied data types and learning tasks.

Our contributions are threefold:\\
(1) We unveil a novel parametrization for learnable augmentation complemented by an adept bilevel algorithm primed for online optimization.\\
(2) Our \SAFLEX method is distinguished by its universal compatibility, allowing it to be effortlessly incorporated into a plethora of supervised learning processes and to collaborate seamlessly with an extensive array of upstream augmentation procedures.\\
(3) The potency of our approach is corroborated by empirical tests on a diverse spectrum of datasets and tasks, all underscoring \SAFLEX's efficiency and versatility, boosting performance by$1.2\%$ on average over all experiments.

\vspace{-0.4em}
\section{Proposed Method: \SAFLEX}
\label{sec:method}
\vspace{-0.4em}

Our goal is to refine augmented samples from any upstream pipeline to enhance classifier generalization. The proposed methodology is founded on two pivotal questions: (1) Which aspects of the augmented samples should be refined? (2) What approach should be taken to learn these refined samples? We start from these questions and defer the derivation of the algorithm to~\cref{sec:algorithm}.

\begin{figure}[htbp!]
\vspace{-1em}
    \centering
    \begin{subfigure}[b]{0.25\textwidth}
    \centering
    \includegraphics[width=1.\linewidth]{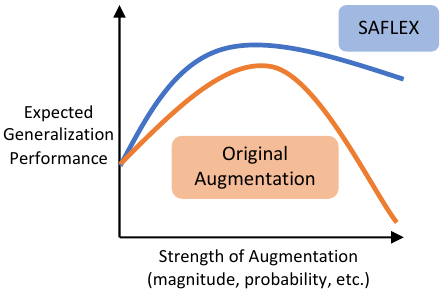}
    \vspace{-1.5em}
    \captionsetup{width=1.1\linewidth}
    \caption{\label{fig:curve}Mitigating trade-offs.}
    \end{subfigure}
    \hspace{20pt}
    \begin{subfigure}[b]{0.46\textwidth}
    \centering
    \includegraphics[width=1.\linewidth]{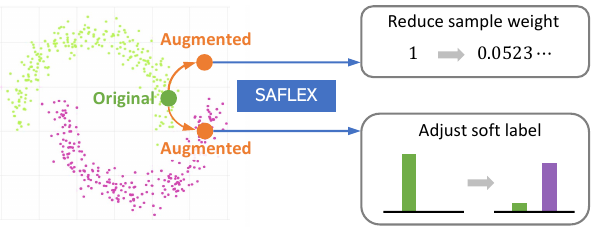}
    \vspace{-1.5em}
    \captionsetup{width=1.1\linewidth}
    \caption{\label{fig:2d-example}Addressing two types of errors.}
    \end{subfigure}
    \vspace{-0.5em}
    \caption{
    \textbf{(a)} Under-augmentation can lead to a scarcity of hard positives, while over-augmentation can introduce an excess of false positives. Reducing the noise in augmentation helps resolve the dilemma. 
    \textbf{(b)} Adjusting sample weights and recalibrating soft labels can address the two types of noises introduced by the augmentation process.}
    \vspace{-1em}
\end{figure}

\cm{Limitations of Augmentation Methods.} Data augmentation is pivotal in enhancing model generalization. 
However, its limitations, particularly the unintentional introduction of noise, can sometimes outweigh its benefits.
For instance, consider the widespread use of random cropping on natural images. Although largely effective, there are times when this approach inadvertently omits task-relevant information, leading to unintended outcomes like false positives. 
This inherent noise creates a trade-off: under-augmentation may yield insufficient challenging examples, whereas over-augmentation can flood the dataset with misleading samples. 
As shown in~\cref{fig:curve}, reducing the noise in augmentation is the key to resolving the dilemma.

Noise in augmentation primarily arises from two fundamental challenges: (1) the deviation of augmented samples from the original data distribution and (2) the potential mislabeling of augmented samples.
We shall envision augmentation as a method to harness prior knowledge in capturing the underlying data distribution. This distribution is encapsulated in the joint distribution, $\mathbb{P}_{XY}(x, y)$, where $x \in \mathcal{X}$ are features and $y \in \{1,\ldots,K\}$ represents labels, with $K$ indicating the number of classes.
Breaking down this joint distribution: $\mathbb{P}_{XY}(x, y) = \mathbb{P}_X(x) \cdot \mathbb{P}_{Y|X} (y|x)$, we observe that the primary source of noise is associated with the feature distribution $\mathbb{P}_X(x)$, while the secondary source is tied to the conditional distribution $\mathbb{P}_{Y|X} (y|x)$.
Addressing these challenges, our methodology is designed to integrate seamlessly with any upstream augmentation process, amending both types of errors post-augmentation, and considering the initial augmentation process as a separate, unchanged entity.

\cm{Feature and Label Extrapolation.}
A key concern in data augmentation pertains to addressing these two types of errors. Some prior works on learning augmentation (e.g., \citep{yang2022adversarial}) attempted to reduce noise by fine-tuning augmented features, using them as initializations. Specifically, the aim was to derive a modified feature $x'$ that eliminates both error types. Yet, due to the high-dimensionality of feature space $\mathcal{X}$, manipulating $x$ is computationally burdensome.

A more efficient strategy is to handle the errors individually and abstain from modifying $x$. When encountering erroneous estimation of the feature distribution $\mathbb{P}_{X}(x)$, even if augmented samples lie in low-density areas, we can compensate by modulating the sample weights $w\in[0,1]$ in the empirical risk minimization loss. Specifically, rarer augmented features are assigned decreased sample weights. For inaccuracies in estimating the conditional distribution $\mathbb{P}_{Y|X} (y|x)$, it's advantageous to modify the augmented label $y$ directly.
We also propose transitioning from a hard class label to a soft one, denoted as $\pl$, representing a probability mass across $K$ classes, residing in the $K$-dimensional simplex $\pl\in\Delta^K$.
The proposed refinement of augmented samples is depicted in~\cref{eq:parametrize}.
Remarkably, optimizing these sample weights and soft labels effectively mitigates errors resulting from varied augmentation methods across numerous classification challenges.
\begin{equation}
    \label{eq:parametrize}
    (x, y) \xrightarrow{\text{Upstream Augment}} (x^\aug, y^\aug) \xrightarrow{\text{SAFLEX}} \stackunder{\big(\quad\sw^\aug}{\text{\footnotesize sample weight $\in[0,1]$}}, x^\aug, \stackunder{\pl^\aug\quad\big)}{\text{\footnotesize soft label $\in\Delta^K$}}
\end{equation}

To elucidate, consider a hypothetical example in \cref{fig:2d-example}. Envision a training sample from the green class (represented by a pronounced green dot). Upon applying a noise-prone augmentation, such as Gaussian perturbation in a 2D setting, the augmented sample could either (1) fall into a region with few validation samples regardless of their class, or (2) be overwhelmingly encompassed by validation samples from a different class. In the former case, it is judicious to reduce the sample weights since they might not be pivotal in discerning the conditional distribution. In the latter instance, the label of the augmented sample should be fine-tuned. This can entail a shift to a soft label to rectify or mitigate potential label inconsistencies, informed by patterns in the validation set.

\cm{Bilevel Formulation.} The remaining question in our design is how to learn the sample weights and soft labels for augmented samples. The overarching goal of augmentation is enhancing model generalization. While the test set remains inaccessible, a prevalent approach is to fine-tune performance using a validation set. This methodology aligns with standard practices in hyperparameter optimization and is evidenced in learnable augmentation methods such as AutoAugment~\citep{cubuk2019autoaugment} and RandAugment~\citep{cubuk2020randaugment}. Given a neural network $\nn{\cdot}:\mathcal{X}\to\Delta^K$ (where we assume Softmax is already applied and the outputs are $\sL_1$ normalized) with parameter $\para$, let us denote the training set, validation set, and the set of augmented samples as $\data_\train$, $\data_\val$, and $\data_\aug$, respectively. The ambition is to refine $\data_\aug$ such that a model trained on the amalgamation of $\data_\train \cup \data_\aug$ optimizes performance on $\data_\val$.
\begin{equation}
    \vspace{-0.5em}
    \label{eq:bilevel}
    \min_{\data_\aug,~\para}\loss(\data_\val, \para) \quad\text{s.t.}\quad \para\in\arg\min_{\para'}\loss(\data_\train\cup\data_\aug, \para')
    \vspace{-0.5em}
\end{equation}
This scenario can be cast as a bilevel optimization problem as in~\cref{eq:bilevel}, where $\data_\aug$, the set of augmented samples with parametrized by sample weights and soft labels, and the model parameters $\para$ are learnable.
The conventional model training constitutes the inner level, while the quest to identify optimal augmented samples $\data_\aug$, which minimize the validation loss post-inner level training, establishes the outer problem. Such a paradigm inherently transforms learnable augmentation into bilevel optimization. Intriguingly, much of the existing literature on learnable augmentation eschews this representation. The primary reservations stem from concerns related to efficiency and differentiability. Notably, works such as \citep{mounsaveng2021learning,mounsaveng2023automatic} are among the sparse few to apply bilevel optimization for augmentation learning, yet their focus remains constricted to affine transformations. In contrast, our approach sidesteps the modification and modeling of feature augmentation, obviating the challenge of differentiability. The low-dimensional nature of sample weights and soft labels potentially simplifies the learning process. In subsequent sections, we demonstrate that, under benign approximations, we can adeptly navigate the bilevel problem, determining the apt sample weights and soft labels within a singular step for each training iteration.
\vspace{-0.4em}
\section{Algorithm}
\label{sec:algorithm}
\vspace{-0.4em}

We now develop an algorithm for the bilevel problem described in~\cref{eq:bilevel}.

\cm{The Greedy Approach.}
Bilevel optimization is notoriously challenging, often necessitating nested loops, which introduces significant computational overhead. Upon inspecting \cref{eq:bilevel}, it becomes evident that an essential characteristic of the problem — the training dynamics of the model — has been understated. In standard practice, augmented samples are typically generated during model training for each minibatch across all iterations.
Therefore, the actual problem deviates from \cref{eq:bilevel} in two ways: (1) different batches of augmentation may influence the learned parameters differently, and the model is not trained on a cumulative set of augmented samples, and conversely, (2) the learned parameters are affected differently by the refined augmented samples across batches, implying that augmentation should be optimized with respect to the corresponding model parameters.

To incorporate model optimization dynamics, we should reformulate the problem on a finer scale: Given the model parameter $\para_{t-1}$ at an intermediate training step, how can we determine the batch of refined augmented samples, $\data^\batch_\aug$? Through a greedy approach, we posit that the granular objective is to minimize the validation loss after a single update, denoted as $\loss(\data_{\val}, \para_t)$, where $\para_t$ is the model parameter updated from $\para_{t-1}$.
\begin{equation}
    \label{eq:greedy}
    \min_{\data^\batch_\aug,~\para_t}\loss(\data_\val, \para_t) \quad\text{s.t.}\quad \para_t=\para_{t-1} - \alpha\cdot\nabla_\para\loss(\data^\batch_\train\cup\data^\batch_\aug, \para_{t-1})
\end{equation}
This micro-perspective of \cref{eq:bilevel} is represented in \cref{eq:greedy}, where the batch of augmented samples $\data^\batch_\aug=\{(\sw^\aug_1, x^\aug_1, \pl^\aug_1), \ldots, (\sw^\aug_B, x^\aug_B, \pl^\aug_B)\}$ is parametrized by the set of sample weights $(\sw^\aug_1,\ldots,\sw^\aug_B)$ and soft labels $(\pl^\aug_1,\ldots,\pl^\aug_B)$. 

As a direct consequence, if the inner loop uses a first-order optimizer like SGD (as assumed), this significantly eases the optimization task. The emergent problem is no longer bilevel. With the analytical solution of the ``inner problem'' at our disposal, we can integrate the formula for $\para_t$ into the outer objective, $\loss(\data_\val, \para_t)$, converting it into a single-level problem.

\cm{Efficient Solution.}
We next derive an algorithm for efficiently addressing~\cref{eq:greedy}. Crucially, due to the linearity of the loss function $\loss(\cdot, \para_{t-1})$ with respect to datasets and the inherent linearity of gradient computation, the gradient vector for the combined training and augmentation batch linearly relates to the sample weights and soft labels, assuming the sample-wise loss function, such as the cross-entropy loss, behaves linearly with respect to sample weights and soft labels. Validating this, the cross-entropy loss is indeed linear concerning these variables, a typical characteristic based on their definitions.

By approximating the validation loss $\loss(\data_\val, \para_t)$ up to the first order around parameter $\theta_t$, we can recast~\cref{eq:greedy} as a linear programming problem.
The objective seeks to maximize the inner product of the gradient vectors on the combined train and augmented batch and the validation batch, effectively yielding a gradient-matching loss~\citep{zhao2020dataset}.
Here, both the objective and normalization constraints linearly correspond to our learnable variables: sample weights and soft labels. The derivation is provided in~\cref{apd:method}, and we summarize the solution in the subsequent notations and theorem.


\begin{notation}
    Let the Jacobian matrix of logits with respect to the model parameter be $\nabla_\theta\nn{x^\aug}\mid_{\theta=\theta_{t-1}}\in\sR^{K\times m}$ and the gradient vector on the validation set be $\nabla_\theta\loss(\data_\val,\theta_{t-1}) \in\sR^{m}$, where $m$ is the parameter count and $K$ is the class count. The Jacobian-vector product is denoted as $\gip=\nabla_\theta\nn{x^\aug}\mid_{\theta=\theta_{t-1}} \nabla_\theta\loss(\data_\val,\theta_{t-1}) \in\sR^K$, which can be computed efficiently.
\end{notation}

\begin{theorem}[Solution of~\cref{eq:greedy}]
\label{thm:saflex}
    The approximated soft label solution is $\pl=\text{OneHot}\left(\arg\max_{k}[\gip]_k\right)$, where $\text{OneHot}(\cdot)$ denotes one-hot encoding, and the sample weight solution is $\sw=1$ if $\sum_{k=1}^K [\gip]_k \geq 0$; otherwise, $\sw=0$.
\vspace{-1em}
\end{theorem}

\cref{thm:saflex} illustrates that an effective approximation of~\cref{eq:greedy} is computationally efficient. The gradient inner product, $\gip$, a Jacobian-vector product, is readily computed alongside standard back-propagation on the combined training and augmentation batch. While determining the validation set gradient vector mandates an additional back-propagation step, we can approximate the gradient vector for the complete validation set, $\nabla_\theta\loss(\data_\val,\theta_{t-1})$, using a minibatch gradient, $\nabla_\theta\loss(\data^\batch_\val,\theta_{t-1})$. Despite necessitating a solution for~\cref{eq:greedy} at every iteration, our efficient \SAFLEX algorithm incurs minimal computational overhead.

A notable takeaway from~\cref{thm:saflex} is that while we aim to learn continuous sample weights (in $[0,1]$) and soft labels (in $\Delta^K$), the derived solutions consistently yield discrete values: either $0$ or $1$ and one-hot vectors. This consistency does not signify a coarse approximation, especially considering we resolve~\cref{eq:greedy} with a $O(\alpha)$ tolerance, where $\alpha$ is typically small. Nonetheless, this characteristic could potentially impact model generalization in under-parameterized scenarios.

\cm{Generalization Aspects.}
Let's interpret and examine the solution provided by \cref{thm:saflex} from a generalization standpoint, which is our primary objective. The loss function's linearity helps understand \cref{thm:saflex}.. Given that \cref{eq:greedy} is a linear program with straightforward normalization constraints, we effectively form a linear combination of $KB$ gradient vectors (each pertaining to a logit of the augmented sample), with $B$ representing the augmented batch size, to approximate the $m$-dimensional validation gradient vector. The total constraints sum up to $B+1$. If these $KB$ gradient vectors are linearly independent, we can always align the combined gradient vector with the validation gradient vector when the degree of freedom, $B + KB - (B+1)$, is greater than or equal to the gradient vector dimension, $m$. This is represented by the condition $KB>m$. Such a scenario, exceedingly under-parametrized, is rare in deep learning. If the combined gradient vector consistently aligns with the validation gradient vector, training with \SAFLEX will approximate training on the combined training and validation sets, potentially limiting the generalization improvements.

To enhance generalization, it is essential to circumvent the challenges of the under-parametrized paradigm, even if we are not closely approaching it. Here, we suggest two modifications to the solution given by \cref{thm:saflex}:

\cm{1. Encouraging Retention of the Original Label.} We can introduce a minor constant penalty term to the gradient inner product to incentivize retaining the augmented sample's original label. Thus, we substitute $\gip$ with $\gip+\beta\mathbf{e}{y^\aug}$, where $\mathbf{e}{y^\aug}$ is a one-hot vector with a value of $1$ at the $y^\aug$-th position. If no other entry in $\gip$ exceeds $[\gip]_{y^\aug}$ by a margin of at least $\beta$, the learned label remains unaltered. This approach proves especially valuable when the validation set is of limited size.

\textbf{2. Substituting $\arg\max$ with Gumbel-SoftMax.} Our current solution invariably yields hard labels. This can sometimes manifest as an excessive degree of confidence, particularly when $\gip$ contains multiple significant entries. To alleviate this, we can employ the Gumbel-SoftMax function to introduce a "softening" effect to the learned labels, adding a measure of stochasticity. Hence, we have $\pl=\mathrm{softmax}\big(\big(\gip+\beta \mathbf{e}_{y^\aug} + \grvv\big)/\tau\big)$, where $\grvv$ consists of i.i.d. random variables sourced from $\text{Gumbel}(0,1)$. Typically, unless specified otherwise, we opt for a relatively low fixed temperature value, $\tau=0.01$.

\cm{The pseudo-code of \SAFLEX} for cross-entropy loss is shown as~\cref{alg:saflex}.

\begin{figure}[htbp!]
\vspace{-1em}
\centering
\begin{minipage}{1.0\textwidth}
\small
\begin{algorithm}[H]
    \caption{\label{alg:saflex}\SAFLEX (Cross-Entropy Loss, Single batch).}
    \DontPrintSemicolon
    \KwIn{Neural network $\nn{\cdot}: \mathcal{X}\to\Delta^{K}$ (softmax applied on outputs) with parameters $\theta$, upstream augmented batch $\{(x^\aug_1, y^\aug_1), \ldots, (x^\aug_B, y^\aug_B)\}$, validation batch $\data^\batch_\val=\{(x^\val_1, y^\val_1), \ldots, (x^\val_{B'}, y^\val_{B'})\}$, penalty coefficient $\beta$, temperature $\tau$.}
    Compute the gradient vector for the validation batch $\nabla_\theta\loss(\data^\batch_\val,\theta)$.\;
    \For(\tcp*[f]{The actual implementation is vectorized.}){$i=1,\ldots,B$}{
        Determine the gradient inner product $\gip_i = \nabla_\theta\nn{x^\aug_i} \nabla_\theta\loss(\data^\batch_\val,\theta)$ via Jacobian-vector product.\;
        Apply Gumbel-SoftMax to get $\pl_i=\mathrm{softmax}\big(\big(\gip_i+\beta \mathbf{e}_{y^\aug_i} + \grvv\big)/\tau\big)$, where $e_{y^\aug_i}\in\sR^{K}$ is one-hot at $y^\aug_i$, and $\grvv$ consists of i.i.d. random variables taken from $\text{Gumbel}(0,1)$.\;
        Set $\sw_i=1$ if $\gip_i\cdot\pl_i\geq 0$, otherwise set $\sw_i=0$.\;
    }
    Renormalize the sample weights $\sw_1,\ldots,\sw_B$ to sum to $1$.\;
    \Return{Sample weights $\sw^\aug_1,\ldots,\sw^\aug_B$, and soft labels $\pl^\aug_1,\ldots,\pl^\aug_B$.}
\end{algorithm}
\end{minipage}
\vspace{-1em}
\end{figure}

\cm{\SAFLEX for Contrastive Learning.}
We conclude this section by discussing to encompass the generalization of the proposed method for certain contrastive learning losses, as illustrated in \cref{eq:contrast} and \cref{eq:clip}. Notably, the latter is utilized for CLIP training. In the realm of contrastive learning, labels are not conventionally defined. Yet, one can perceive the contrastive training objectives in \cref{eq:contrast} and \cref{eq:clip} as proxy classification tasks. Here, we posit that the batch of size $B$ can be construed as containing $B$ classes: one positive example coupled with $B-1$ negative examples. This interpretation paves the way to introduce the notion of (soft) labels over this surrogate classification task with its $B$ distinct classes.

Under this paradigm, the loss function remains linear concerning the soft labels and sample weights, making the methodology in \cref{thm:saflex} applicable. The sole requisite modification pertains to the gradient inner product's definition. Rather than employing gradients from the cross-entropy logits, $\nabla_\para\nn{x^\aug}$, we substitute them with gradients corresponding to the contrastive learning logits.

\vspace{-0.4em}
\section{Related Works}
\label{sec:related}
\vspace{-0.4em}

Traditional data augmentation techniques such as random flipping and cropping~\citep{krizhevsky2017imagenet,simard2003best,shorten2019survey} are hand-crafted and static, unlike our adaptive \SAFLEX method that tunes sample weights based on validation performance. Autonomous approaches like AutoAugment~\citep{cubuk2019autoaugment,lim2019fast,ho2019population,mounsaveng2021learning,mounsaveng2023automatic} learn transformations but are restricted in scope, primarily focusing on affine transformations. Generative methods employing GANs or diffusion models~\citep{odena2017conditional,sankaranarayanan2018generate,huang2018auggan,he2022synthetic,shipard2023diversity,dunlap2023diversify,trabucco2023effective} can inadvertently alter class-relevant features, which our method avoids by adaptively adjusting sample weights. Research on adversarial perturbations~\citep{goodfellow2015explaining,yang2022adversarial,yang2022identity,ho2020contrastive} and noise-robust learning~\citep{han2018co,lang2022training,thulasidasan2019combating,konstantinov2019robust,gao2022self,ma2018dimensionality,kremer2018robust} address similar problems but often suffer from complexity and stability issues, which we mitigate by our principled approach to weight adjustment.
Recently, Soft-Augmentation~\citep{liu2023soft} also proposes to use soft labels and sample weights to train on augmented samples. However, it implements a specific formula to generate them based on the strength parameter of upstream augmentations. This limits the applicability of Soft-Augmentation mostly to crop augmentation on images.
\citet{wang2023metamix} introduce self-adaptive augmentation within the meta-learning framework, MetaMix, which improves the corruption robustness of continual learning models.
\citet{bhattarai2020sampling} propose a progressive sampling strategy for GAN synthetic data, while \citet{caramalau2021sequential} introduce a sequential graph convolutional network for active learning.
Our work extends these findings by developing a novel sampling and purifying method for augmented data that is specifically designed to improve the performance of downstream tasks.

For a more detailed discussion of related works, please refer to \cref{sec:related_apd}.
\vspace{-0.4em}
\section{Experiments}
\label{sec:experiment}
\vspace{-0.4em}

We validate the effectiveness of \SAFLEX under four very different learning scenarios: (1) adapting augmentations to medical images, (2) refining augmentations for tabular data, (3) purifying diffusion-model-generated augments, and (4) applying to contrastive fine-tuning. Experimental setups and implementation details are provided in Appendix~\ref{apd:implement}.

\cm{Adapting Augmentations to Medical Images.}
Unlike natural images, medical images often carry quantitative information (e.g., encoded as color) and objects without a canonical orientation. While we usually lack the domain knowledge to design effective heuristic augmentation transformations for these images, applying augmentation pipelines designed for natural images, such as RandAugment~\citep{cubuk2020randaugment}, can sometimes degrade performance in the medical context~\citep{yang2022adversarial}. Consequently, we investigate whether \SAFLEX can adapt these augmentation pipelines for medical images.

We assess multi-class classification across eight medical image datasets from MedMNIST~\citep{yang2023medmnist}, with each dataset comprising 10K to 236K 28$\times$28 images and 4 to 11 classes. In line with~\citep{yang2021medmnist}, we train a ResNet-18 model~\citep{he2016deep} using the Adam optimizer~\citep{kingma2014adam} for 100 epochs. For upstream augmentation, we utilize the widely-adopted RandAugment~\citep{cubuk2020randaugment} and Mixup~\citep{zhang2018mixup} methods. Test accuracies are presented in \cref{tab:medimg}, highlighting that \SAFLEX significantly enhances the performance of both RandAugment and Mixup.
It's noteworthy that \SAFLEX, when combined with basic upstream augmentations as shown in~\cref{tab:medimg}, achieves better performance than Soft-Augmentation~\citep{liu2023soft}, and comparable or superior performance than the adversarial-perturbation-based augmentation, LP-A3~\citep{yang2022adversarial}. The latter not only takes significantly longer to train but also demands careful hyperparameter tuning. For a comprehensive view, Soft-Augmentation's performance and LP-A3's performance on the MedMNIST datasets can be found in~\cref{apd:implement}.

\begin{table}[htbp!]
\centering
\adjustbox{max width=1.00\textwidth}{%
{\renewcommand{\arraystretch}{1.20}%
{\Large
\centering
\begin{tabular}{lcccccccc}
    \toprule
    \textbf{Method} & \textbf{Path} & \textbf{Derma} & \textbf{Tissue} & \textbf{Blood} & \textbf{OCT} & \textbf{OrganA} & \textbf{OrganC} & \textbf{OrganS} \\
    \midrule
    No Aug & $94.34 \pm 0.18$ & $76.14 \pm 0.09$ & $68.28 \pm 0.17$ & $96.81 \pm 0.19$ & $78.67 \pm 0.26$ & $94.21 \pm 0.09$ & $91.81 \pm 0.12$ & $81.57 \pm 0.07$ \\
    RandAug & $93.52 \pm 0.09$ & $73.71 \pm 0.33$ & $62.03 \pm 0.14$ & $95.00 \pm 0.21$ & $76.00 \pm 0.24$ & $94.18 \pm 0.20$ & $91.38 \pm 0.14$ & $80.52 \pm 0.32$ \\
    \rowcolor{lightgray} \begin{tabular}[c]{@{}l@{}}\rowcolor{lightgray}\SAFLEX\\ (w/ RandAug)\end{tabular} & \pmb{$95.11 \pm 0.14$} & $76.69 \pm 0.33$ & $64.32 \pm 0.18$ & $96.91 \pm 0.15$ & \pmb{$79.63 \pm 0.28$} & \pmb{$95.32 \pm 0.29$} & $92.10 \pm 0.21$ & \pmb{$82.85 \pm 0.42$} \\
    Mixup & $92.98 \pm 0.19$ & $75.22 \pm 0.45$ & $66.62 \pm 0.31$ & $96.28 \pm 0.23$ & $77.93 \pm 0.41$ & $94.12 \pm 0.35$ & $90.76 \pm 0.28$ & $80.99 \pm 0.21$ \\
    \rowcolor{lightgray} \begin{tabular}[c]{@{}l@{}}\rowcolor{lightgray}\SAFLEX\\ (w/ Mixup)\end{tabular} & $93.71 \pm 0.37$ & \pmb{$76.94 \pm 0.51$} & \pmb{$68.31 \pm 0.43$} & \pmb{$97.21 \pm 0.35$} & $79.54 \pm 0.44$ & $95.06 \pm 0.31$ & \pmb{$92.73 \pm 0.53$} & $82.14 \pm 0.27$ \\
    \bottomrule
\end{tabular}
}}}
    \caption{On \textbf{medical images}, \SAFLEX significantly enhances the performance of RandAugment and Mixup across eight medical image datasets from MedMNIST.}
    \label{tab:medimg}
\end{table}

In terms of efficiency, \SAFLEX, designed as an augmentation plug-in, requires only a single-step update per iteration. It only extends the average wall-clock time of a training epoch by roughly 42\% in this experiment; see~\cref{apd:implement} for details.

\cm{Refining Augmentations for Tabular Data.}
Tabular data typically encompasses heterogeneous features that include a blend of continuous, categorical, and ordinal values. The presence of discrete features constrains the space of potential transformations. Furthermore, the domain knowledge to design invariant, label-preserving transformations is often absent. One of the few traditional augmentation techniques directly applicable to tabular data is CutMix~\citep{yun2019cutmix}, which substitutes a portion of continuous or discrete features with values from other randomly chosen samples (see~\cref{apd:implement} for implementation details). However, studies suggest that CutMix, with a relatively small augmentation probability like $0.1$, struggles to bolster tabular classification performance~\citep{onishi2023rethinking}. Conversely, a higher augmentation probability can introduce excessive noise, potentially downgrading the performance. This leads us to explore whether \SAFLEX can mitigate the noise from CutMix and enhance classification performance.

Our experiments span seven tabular datasets varying in size (from 452 to 494K) and feature types (from exclusively continuous features to predominantly discrete ones); detailed dataset information and statistics are available in~\cref{apd:implement}. Except for the \textit{Volkert} dataset, which involves 10-way classification, all other datasets focus on binary classification. Notably, some datasets, like \textit{Credit}, exhibit a significantly skewed class distribution (e.g., only 0.17\% positive). We consider backbone models such as the sample Multilayer Perceptron (MLP) with two hidden layers and 256 neurons each and tranformer-based models like SAINT~\citep{somepalli2022saint} (without contrastive pretraining). These models undergo training with dropout~\citep{srivastava2014dropout} and, in certain cases, batch normalization, for 200 epochs.

\begin{table}[htbp!]
\centering
\adjustbox{max width=1.00\textwidth}{%
{\renewcommand{\arraystretch}{1.10}%
{\Large
\centering
    \begin{tabular}{llccccccc}
    \toprule
    \textbf{Method} & \textbf{Model} & \textbf{Appetency} & \textbf{Arrhythmia} & \textbf{Click} & \textbf{Credit} & \textbf{QASR} & \textbf{Shrutime} & \textbf{Volkert} \\
    \midrule
    No Aug & MLP & $49.03 \pm 0.01$ & $81.53 \pm 0.03$ & $52.54 \pm 0.04$ & $66.91 \pm 0.03$ & $91.84 \pm 0.02$ & $86.27 \pm 0.04$ & $61.14 \pm 0.05$ \\
    CutMix & MLP & $48.98 \pm 0.03$ & $81.57 \pm 0.05$ & $52.59 \pm 0.09$ & $73.68 \pm 0.08$ & $91.87 \pm 0.02$ & $86.39 \pm 0.05$ & $61.20 \pm 0.02$ \\
    \rowcolor{lightgray} \begin{tabular}[c]{@{}l@{}}\rowcolor{lightgray}\SAFLEX\\ (w/ CutMix)\end{tabular} & MLP & \pmb{$51.04 \pm 0.09$} & \pmb{$83.02 \pm 0.06$} & \pmb{$52.81 \pm 0.06$} & \pmb{$74.61 \pm 0.15$} & \pmb{$92.69 \pm 0.13$} & \pmb{$86.90 \pm 0.10$} & \pmb{$61.51 \pm 0.05$} \\ \hline
    No Aug & SAINT & $78.90 \pm 0.03$ & $83.90 \pm 0.01$ & $65.72 \pm 0.06$ & $79.49 \pm 0.05$ & $98.18 \pm 0.04$ & $87.53 \pm 0.04$ & $66.82 \pm 0.05$ \\
    CutMix & SAINT & $81.05 \pm 0.07$ & \pmb{$85.32 \pm 0.09$} & $65.77 \pm 0.04$ & $79.71 \pm 0.08$ & \pmb{$98.61 \pm 0.06$} & $87.61 \pm 0.07$ & $68.23 \pm 0.10$ \\
    \rowcolor{lightgray} \begin{tabular}[c]{@{}l@{}}\rowcolor{lightgray}\SAFLEX\\ (w/ CutMix)\end{tabular} & SAINT & \pmb{$81.33 \pm 0.14$} & $85.27 \pm 0.14$ & \pmb{$66.12 \pm 0.09$} & \pmb{$79.93 \pm 0.17$} & $98.59 \pm 0.21$ & \pmb{$87.93 \pm 0.13$} & \pmb{$68.91 \pm 0.17$} \\
    \bottomrule
    \end{tabular}
    }}}
    \caption{On \textbf{tabular data}, \SAFLEX outperforms the upstream augmentation method, CutMix, across diverse tabular datasets using either MLP or SAINT as the backbone models.}
    \label{tab:tabular}
\end{table}

\cref{tab:tabular} shows that \SAFLEX almost consistently enhances the performance of CutMix across all datasets, regardless of whether the MLP or SAINT model is used. This improvement is especially noticeable with the MLP backbone, which is typically more intricate to train, and on datasets abundant in discrete features, such as \textit{Click} and \textit{Shrutime}, where CutMix tends to inject more noise. Notably, the \textit{Volkert} dataset demonstrates a considerable performance impovement, potentially attributed to the fact that it has 10 classes where soft labels might be more useful.

\cm{Purifying Diffusion-Model-Generated Augments.}
Recent research \citep{dunlap2023diversify,trabucco2023effective} has advocated the application of diffusion models for image editing via text prompts. Compared to traditional augmentation techniques, images produced by pretrained diffusion models maintain task-specific details while offering enhanced domain diversity, as dictated by the prompts. Diffusion-model-generated augmentations have been found particularly efficacious in fine-grained classification and out-of-distribution (OOD) generalization tasks~\citep{dunlap2023diversify}. However, these diffusion models occasionally generate subtle image alterations, potentially corrupting class-essential information, thus underscoring the necessity for noise reduction~\citep{dunlap2023diversify}. In this context, we probe the capability of \SAFLEX to enhance the purity of diffusion-model-generated augmentations, aiming for improved classification outcomes.

In our experimentation, we adhere to the setups in~\citep{dunlap2023diversify}. We assess \SAFLEX using diffusion-model-generated images derived from two distinct approaches:
(1) The \textit{Img2Img} approach involves an image encoder that first converts a given image into a latent representation. Subsequently, employing a diffusion model (specifically, Stable Diffusion v1.5~\citep{rombach2022high} for this experiment), this latent representation undergoes a series of prompt-conditioned transformations. Ultimately, the altered representation is decoded, yielding an augmented image reflecting the modifications stipulated in the prompt. Notably, the diffusion model may or may not undergo fine-tuning (\textit{w/ and w/o finetune}) on the dataset in question.
(2) The \textit{InstructPix2Pix} approach~\citep{brooks2023instructpix2pix} accepts an image and an edit instruction sampled (e.g., “position the animals within the forest”) and outputs a correspondingly modified image. InstructPix2Pix is a conditional diffusion model pretrained on a dataset containing paired images and their associated edit instructions.

Our evaluation encompasses two tasks:
(1) Fine-grained classification on a CUB dataset subset~\citep{wah2011caltech} (featuring 25 images per category).
(2) OOD generalization on an iWildCam subset from the Wilds dataset~\citep{koh2021wilds} (consisting of over 6,000 images and simplified to 7-way classification).
We use a ResNet-50 model~\citep{he2016deep} pretrained on ImageNet~\citep{deng2009imagenet}.
For comparison, we also consider data generated solely from text (\textit{Text2Img}) and the \textit{RandAugment} method as baselines.

\begin{table}[htbp!]
\centering
\adjustbox{max width=1.00\textwidth}{%
{\renewcommand{\arraystretch}{1.25}%
{\Huge
\centering
\begin{tabular}{lcccc>{\columncolor{lightgray}}cc>{\columncolor{lightgray}}cc>{\columncolor{lightgray}}c}
    \toprule
    \textbf{Task} & \textbf{No Aug} & \textbf{RandAug} & \textbf{Text2Img} & \multicolumn{2}{c}{\textbf{InstructPix2Pix}} & \multicolumn{2}{c}{\textbf{Img2Img (w/o finetune)}} & \multicolumn{2}{c}{\textbf{Img2Img (w/ finetune)}} \\
    \midrule
     & --- & --- & --- & {\huge w/o \SAFLEX} & {\huge w/ \SAFLEX} & {\huge w/o \SAFLEX} & {\huge w/ \SAFLEX} & {\huge w/o \SAFLEX} & {\huge w/ \SAFLEX} \\
    \begin{tabular}[c]{@{}l@{}}Fine-Grained\\ Classification\end{tabular} & $68.60 \pm 0.16$ & $71.26 \pm 0.52$ & $69.68 \pm 0.97$ & $71.38 \pm 0.91$ & \pmb{$72.34 \pm 0.59$} & $71.25 \pm 0.86$ & \pmb{$73.22 \pm 0.63$} & $72.01 \pm 1.24$ & \pmb{$73.61 \pm 0.78$} \\
    \midrule
    \begin{tabular}[c]{@{}l@{}}OOD\\ Generalization\end{tabular} & $57.19 \pm 1.13$ & $61.34 \pm 2.72$ & $64.53 \pm 3.01$ & $67.29 \pm 1.96$ & \pmb{$69.92 \pm 0.88$} & $70.65 \pm 1.50$ & \pmb{$72.61 \pm 1.44$} & $70.49 \pm 1.21$ & \pmb{$72.83 \pm 0.92$} \\
    \bottomrule
\end{tabular}
}}}
\caption{For \textbf{diffusion-model-generated augmentations}, \SAFLEX enhances the fine-grained classification and OOD generalization performance for various diffusion-generation methods.}
\label{tab:diffusion}
\end{table}

Results, as depicted in~\cref{tab:diffusion}, affirm that \SAFLEX consistently elevates the performance of all three diffusion-model-generated augmentation techniques, across both fine-grained classification and OOD generalization tasks. Notably, the performance boost is more prominent within the OOD generalization task, where feature and label distortions are particularly detrimental. We confirm that \SAFLEX is useful to refine diffusion-model-generated augmentations, leading to enhanced classification accuracy. 

\cm{Applying to Contrastive Fine-Tuning.}
We next shift our focus from the empirical risk minimization (ERM) framework utilizing cross-entropy loss, as demonstrated in the prior scenarios. To test the adaptability and compatibility of \SAFLEX with contrastive loss, we turn to a contrastive fine-tuning paradigm termed ``Finetune Like You Pretrain'' (FLYP)\citep{goyal2023finetune}. This methodology offers a straightforward yet potent means to fine-tune pretrained image-text models, including notable ones like CLIP~\citep{radford2021learning}. Remarkably, by simply fine-tuning classifiers through the initial pretraining contrastive loss (refer to \cref{eq:clip}), FLYP achieves uniformly better classification performance. This entails constructing prompts from class labels and subsequently minimizing the contrastive loss between these prompts and the image embeddings within the fine-tuning set.

Our experimentation adopts the framework presented in~\citep{goyal2023finetune}. Specifically, we fine-tune a CLIP model equipped with a ViT-B/16 encoder on the full iWildCam dataset from Wilds~\citep{koh2021wilds}. Post fine-tuning, we adopt a strategy from~\citep{goyal2023finetune} that linearly interpolates weights between the pretrained and the fine-tuned checkpoints to optimize in-distribution (ID) performance. As our upstream augmentation technique, we employ RandAugment, following hyperparameter setups as described in~\citep{koh2021wilds}. For handling the CLIP contrastive loss, we apply our tailored algorithm, detailed in~\cref{sec:algorithm}. For an in-depth understanding, please refer to~\cref{apd:method} and~\cref{apd:implement}.

\begin{table}[htbp!]
    \centering
    \adjustbox{max width=0.9\textwidth}{%
    {\renewcommand{\arraystretch}{1.15}%
    {\Huge
    \centering
    \begin{tabular}{lccccc>{\columncolor{lightgray}}c}
        \toprule
        \textbf{Task} &  & \textbf{Zero-Shot} & \textbf{LP-FT} & \textbf{FLYP} & \textbf{FLYP+RandAug} & \textbf{FLYP+\SAFLEX(w/ RandAug)} \\
        \midrule
        ID & \multirow{2}{*}{\huge w/o Ensembling} & $8.7 \pm 0.0$ & $49.7 \pm 0.5$ & $52.2 \pm 0.6$ & $52.4 \pm 0.8$ & \pmb{$52.7 \pm 0.7$} \\
        OOD &  & $11.0 \pm 0.0$ & $34.7 \pm 0.4$ & $35.6 \pm 1.2$ & $36.3 \pm 1.4$ & \pmb{$36.9 \pm 1.5$} \\
        \midrule
        ID & \multirow{2}{*}{\huge w/ Ensembling} & $8.7 \pm 0.0$ & $50.2 \pm 0.5$ & $52.5 \pm 0.6$ & $52.6 \pm 1.0$ & \pmb{$53.0 \pm 0.7$} \\
        OOD &  & $11.0 \pm 0.0$ & $35.7 \pm 0.4$ & $37.1 \pm 1.2$ & $37.6 \pm 0.9$ & \pmb{$37.8 \pm 1.1$} \\
        \bottomrule
    \end{tabular}
    }}}
    \caption{Applied to \textbf{contrastive fine-tuning of CLIP using FLYP}~\citep{goyal2023finetune}, \SAFLEX also enhances the performance of standard image augmentations like RandAugment.}
    \label{tab:FLYP}
\end{table}

As evidenced in~\cref{tab:FLYP}, incorporating RandAugment alongside FLYP yields favorable outcomes. Moreover, the introduction of \SAFLEX amplifies performance gains for both ID and OOD tasks, irrespective of whether ensembling is applied. This observation is particularly noteworthy, as it demonstrates that \SAFLEX is compatible with contrastive loss, which is a key component of many training paradigms, including self-supervised learning.

\vspace{-0.4em}
\section{Conclusions}
\label{sec:conclusion}
\vspace{-0.4em}

Our study presents \SAFLEX, a novel solution to current challenges in data augmentation.
At its core, \SAFLEX offers a paradigm shift from traditional, one-size-fits-all augmentation strategies to a more adaptive, data-driven approach. 
It allows for the learning of low-dimensional sample weights and soft labels for each augmented instance, thereby circumventing the complexities and limitations inherent in direct augmentation feature learning. 
Our method demonstrates universal compatibility, underscoring its vast potential for diverse data types in learning scenarios. 
Extensive empirical evaluations confirm \SAFLEX's prowess, proving its adaptability from medical imaging contexts to the nuances of tabular and natural image datasets.
While \SAFLEX demonstrates promising results, there are certain factors to consider for optimal performance. A substantial and high-quality validation set is beneficial. A suboptimal set could limit its effectiveness. Additionally, the type of upstream augmentation methods selected plays a role, as it impacts the overall performance of \SAFLEX. Our approach also entails some computational overhead due to frequent gradient evaluations. These considerations will be the focus of future studies to further refine the methodology.
In essence, \SAFLEX stands as a testament to the advancements in learnable data augmentation, ushering in a more adaptive and customized era of data-centric AI.

\clearpage
\newpage

\subsubsection*{Acknowledgments}
Ding, An, Xu, Satheesh, and Huang are supported by National Science Foundation NSF-IIS-2147276 FAI, DOD-ONR-Office of Naval Research under award number N00014-22-1-2335, DOD-AFOSR-Air Force Office of Scientific Research under award number FA9550-23-1-0048, DOD-DARPA-Defense Advanced Research Projects Agency Guaranteeing AI Robustness against Deception (GARD) HR00112020007, Adobe, Capital One and JP Morgan faculty fellowships.

\bibliography{references}
\bibliographystyle{iclr2024_conference}

\clearpage
\newpage
\appendix
\section{Method and Algorithm Details}
\label{apd:method}

In this appendix, we provide more details about the proposed method and algorithm.
We first show the derivation details behind results in~\cref{thm:saflex}.
Then, we provide more details about the \SAFLEX algorithm on contrastive losses discussed at the end of~\cref{sec:algorithm}.

\begin{prevproof}{thm:saflex}
    As outlined in~\cref{sec:algorithm}, we start from approximating the validation loss up to first order around the current parameter $\para_{t-1}$.
    By the first-order approximation, we shall rewrite the optimization problem in~\cref{eq:greedy} as follows:
    \begin{equation}
        \label{eq:approximated}
        \max_{\substack{(\sw_1,\ldots,\sw_B),~(\pl_1,\ldots,\pl_B);\\ \sum_{i=1}^B\sw_i=1,~\pl_i\in\Delta^K, \forall i\in[B]}}\Big\langle\nabla_\para\loss(\data_\val, \para_{t-1}), \nabla_\para\loss\big(\data^\batch_\train\cup\{(\sw^\aug_i, x^\aug_i, \pl^\aug_i)\}_{i=1}^B, \para_{t-1}\big)\Big\rangle
    \end{equation}
    where we also explicitly write out the learnable parts in the augmented batch.

    Clearly, the set of constraints, $\sum_{i=1}^B\sw_i=1$ and $
    \sum_{k=1} [K\pl_i]_k=1$ for $\forall i\in[B]$, are linear.
    To show that the objective function is also linear, we consider the form of cross-entropy loss:
    \begin{equation}
        \label{eq:ce}
        \loss_\ce(\data, \para) = -\sum_{i=1}^{B} \log \frac{\exp\left[\nn{x_i}\right]_{y_i}}{\sum_{k=1}^K\exp\left[\nn{x_i}\right]_k}
    \end{equation}
    Since $\nn{\cdot}:\mathcal{X}\to\Delta^K$ is assumed to have the Softmax function applied on the outputs (see~\cref{sec:method}), we have $\sum_{k=1}^K\exp\left[\nn{x_i}\right]_k=1$, and the cross-entropy loss can be rewritten as:
    \begin{equation}
        \loss_\ce(\data, \para) = -\sum_{i=1}^{B} \log [\nn{x_i}]_{y_i}
    \end{equation}
    When sample weights and soft labels are introduced, the cross-entropy loss becomes:
    \begin{equation}
        \loss_\ce(\data, \para) = -\sum_{i=1}^{B} \sw_i \cdot \sum_{k=1}^K [\pl_i]_k \log [\nn{x_i}]_k
    \end{equation}
    From the above equation, we can see that the objective function $\loss\big(\data^\batch_\train\cup\{(\sw^\aug_i, x^\aug_i, \pl^\aug_i)\}_{i=1}^B, \para_{t-1}\big)$ in~\cref{eq:approximated} is indeed linear with respect to sample weigths $(\sw_1,\ldots,\sw_B)$ and soft labels $(\pl_1,\ldots,\pl_B)$.

    Given these, we conclude, the resulted optimizaiton task,~\cref{eq:approximated}, is a linear programming problem, which can be solved efficiently.
    Moreover, the set of linear constraints are independent, which means the solution for sample weight $\sw$ and soft labels $\pl$ for an augmented sample $x^\aug\in\data^\batch_\aug$ are independent of other augmented samples and the training sample batch $\data^\batch_\train$.
    For an arbitrary augmented sample $x^\aug\in\data^\batch_\aug$, replacing the gradient vector on the entire batch of training and augmented samples with the gradient vector on this single augmented sample,
    \begin{equation}
        \loss_\ce(\{(\sw^\aug, x^\aug, \pl^\aug)\}, \para) = \sw^\aug \cdot \sum_{k=1}^K [\pl^\aug]_k \log [\nn{x^\aug}]_k
    \end{equation}
    it is not hard to see that if the gradient inner product is denoted by
    \begin{equation}
        \gip=\nabla_\theta\nn{x^\aug}\mid_{\theta=\theta_{t-1}} \nabla_\theta\loss(\data_\val,\theta_{t-1})
    \end{equation}
    the optimal solution for $\sw^\aug$ and $\pl^\aug$ are:
    \begin{equation}
        \pl=\text{OneHot}\left(\arg\max_{k}[\gip]_k\right)
    \end{equation}
    where $\text{OneHot}(\cdot)$ denotes one-hot encoding, and,
    \begin{equation}
        \sw=1 \text{ if } \sum_{k=1}^K [\gip]_k \geq 0 \text{,~~otherwise } \sw=0
    \end{equation}
\end{prevproof}

For the adaptation of \SAFLEX to contrastive losses, in~\cref{sec:algorithm}, we have already shown the main idea.
Here we take a closer look at some typical contrastive losses like,
\begin{equation}
    \label{eq:contrast}
    \loss_\contrast(\data, \para) = -\sum_{i=1}^{B} \log \frac{\exp\left(\cenn{x_i} \cdot \cenn{x^+_i}/ \tau\right)}{\exp\left(\cenn{x_i} \cdot \cenn{x^+_i} / \tau\right) + \sum_{j=1}^{B-1} \exp \left(\cenn{x_i} \cdot \cenn{x^-_j} / \tau\right)}
\end{equation}
where $x^+_i$ is the positive example of $x_i$, and $x^-_j$ is the $j$-th negative example of $x_i$. And $\cenn{\cdot}$ is the $\mathcal{L}_2$ normalized encoder, $\tau$ is the temperature.
And the contrastive pre-training (which is also use for finetuning in~\citep{goyal2023finetune}),
\begin{equation}
    \label{eq:clip}
    \loss_\clip(\data, \para):= \sum_{i=1}^B-\log \frac{\exp \left(\ienn{I_i} \cdot \tenn{T_i}\right)}{\sum_{j=1}^B \exp \left(\ienn{I_i} \cdot \tenn{T_j}\right)}+ \sum_{i=1}^B-\log \frac{\exp \left(\ienn{I_i} \cdot \tenn{T_i}\right)}{\sum_{j=1}^B \exp \left(\ienn{I_j} \cdot \tenn{T_i}\right)},
\end{equation}
where $I_i$ is the image, and $T_i$ is the text for the $i$-th sample. $\ienn{\cdot}$ and $\tenn{\cdot}$ are the $\mathcal{L}_2$ normalized image and text encoders, respectively.

We confirm that one can perceive the contrastive training objectives in \cref{eq:contrast} and \cref{eq:clip} as proxy classification tasks.
Here, we posit that the batch of size $B$ can be construed as containing $B$ classes: one positive example coupled with $B-1$ negative examples.
This interpretation paves the way to introduce the notion of (soft) labels over this surrogate classification task with its $B$ distinct classes.

Taking the CLIP loss as an example, we shall generalize the first term in \cref{eq:clip} to the following:
\begin{equation}
    \sum_{i=1}^B -\sw_i \cdot \sum_{j=1}^B [\pl_i]_j \log \frac{\exp \left(\ienn{I_i} \cdot \tenn{T_j}\right)}{\sum_{k=1}^B \exp \left(\ienn{I_i} \cdot \tenn{T_j}\right)}
\end{equation}
where there are $B$ proxy-classes.

Under this paradigm, the loss function remains linear concerning the soft labels and sample weights, making the methodology in \cref{thm:saflex} applicable.

\section{Related Work}
\label{sec:related_apd}

In this section, we compare our approach with established augmentation methods, including traditional heuristical transformations, autonomous data augmentation, and methods leveraging large pretrained models or adversarial perturbation. We then discuss our methodology's connections to noise-robust learning and hyperparameter optimization. For a detailed background on our experimental tasks and other connected areas.

\cm{Traditional data augmentation} operations are usually crafted and chosen heuristically based on domain expertise~\citep{krizhevsky2017imagenet,simard2003best}. For natural images, common transformations include random flipping, cropping, and color shifting~\citep{shorten2019survey}. Mixup-based~\citep{zhang2018mixup} augmentations like cutmix~\citep{yun2019cutmix} enhance data diversity by merging patches from two images, which is also widely adopted for tabular datasets. Although we, like mixup, introduce soft labels, ours are not the outcome of merging two data instances. Nevertheless, traditional methods enjoy no guarantee of effectiveness or universality, limiting their applicability across varied data types and tasks.

\cm{Autonomous data augmentation} has a rich history, while classical works generally bifurcate into AutoAugment-based and GAN-based approaches. \cm{AutoAugment}~\citep{cubuk2019autoaugment} learns sequences of transformations to optimize classifier performance on a validation set. Subsequent works~\citep{lim2019fast,ho2019population} have proposed alternative learning algorithms. Among them, \citep{mounsaveng2021learning,mounsaveng2023automatic} propose to learn the augmentation transformation using bilevel optimization at the cost that only differentiable affine transformations can be considered. Subsequently, RandAugment~\citep{cubuk2020randaugment} demonstrates equivalent performance to AutoAugment by employing random transformation selection. However, such approaches still rely on a priori knowledge of beneficial transformations. On the other hand, \cm{GAN-generated} images conditioned on their class can be used as augmented samples~\citep{odena2017conditional,sankaranarayanan2018generate,huang2018auggan}. However, the inherent assumption of GANs, that augmented data should mimic the original distribution, often restricts potential enhancements~\citep{shorten2019survey}. 

\cm{Pretrained large generative models}, like \cm{diffusion models}, offer the capability to synthesize training data in zero or few shot scenarios~\citep{he2022synthetic,shipard2023diversity} as well as generate hard training examples~\citep{jain2022distilling}.
Nonetheless, models exclusively trained on diffusion-produced data often underperform compared to their counterparts trained on real datasets~\citep{azizi2023synthetic}. To address this, recent studies~\citep{dunlap2023diversify,trabucco2023effective} proposed the use of diffusion models for image editing with text prompts, yielding augmentations closer to original training data without necessitating finetuning. In contrast to conventional GAN-based methods, \cm{diffusion-based augmentations} leverage knowledge from large pretrained datasets. However, they can sometimes produce subtle image edits and corrupt class-relevant information, highlighting the importance of noise reduction techniques like filtering~\citep{dunlap2023diversify}. While such filtering relies on heuristic metrics, it can be viewed as a specific case of learning sample weights in our work. Another line of research models augmentation as \cm{adversarial perturbations}~\citep{goodfellow2015explaining}, aiming to generate more challenging positive and negative samples~\citep{yang2022adversarial,yang2022identity,ho2020contrastive}. However, these models usually suffer from inherent complexity and instability issues.

\cm{Noise robust learning} bears relevance to our approach since we treat upstream augmented samples as noisy data.
Learning sample weights and soft labels parallel noise reduction strategies such as dataset resampling~\citep{han2018co,lang2022training}, loss reweighting~\citep{thulasidasan2019combating,konstantinov2019robust,gao2022self}, and label correction~\citep{ma2018dimensionality,kremer2018robust}. Our method is efficient yet principled as we formulate to optimize the model performance on the validation set, similar to standard \cm{hyperparameter search} paradigms. Our algorithm bears relevance to continuous hyperparameter optimization~\cite{lorraine2020optimizing} in its use of bilevel optimization algorithms~\citep{liu2022bome}, but we introduce a novel bilevel approach. Data augmentation is greedily learned in our formulation, in sync with the ongoing training dynamics.

\cm{Medical image classification}
MedMNIST~\citep{yang2023medmnist,yang2021medmnist} is a comprehensive dataset of biomedical images, offering both 2D and 3D standardized images pre-processed to small sizes with classification labels.
ResNets~\citep{he2016deep} popular models for medical image classification.
\citep{yang2022adversarial} and \citep{mounsaveng2023automatic} are augmentation methods that have been shown to improve performance on medical image classification tasks.
\citep{yang2022adversarial} introduces a novel, prior-free autonomous data augmentation approach that leverages representation learning to create hard positive examples as augmentations, enhancing performance in various machine learning tasks without the need for a separate generative model.
\citep{mounsaveng2023automatic} proposes an automatic data augmentation learning method for histopathological images, wherein the augmentation parameters are determined as learnable using a bilevel optimization approach, proving more efficient and effective than predefined transformations. However, \citep{mounsaveng2023automatic} is not evaluated on MedMNIST and the adaptation is non-trivial.

\cm{Tabular data classification}
Classical models, such as XGBoost~\citep{chen2016xgboost}, have been the cornerstone for tabular data processing, providing interpretability and handling diverse feature types effectively, including those with missing values.
Multilayer perceptrons (MLPs) have also been a staple in the domain, offering flexibility in modeling non-linear relationships in tabular datasets. 
TabNet~\citep{arik2021tabnet}, a more recent innovation, employs neural networks to emulate decision trees, focusing selectively on specific features at every layer. 
Lastly, SAINT~\citep{somepalli2022saint} presents a hybrid deep learning solution tailored for tabular data. It employs attention mechanisms over both rows and columns and introduces an improved embedding technique.
On tabular data, cutmix~\citep{yun2019cutmix} is widely adopted and considered as a standard augmentation method.

\cm{Diffusion-model-based image augmentations}
Recent studies have shed light on the prowess of diffusion models in image augmentations. \citep{dunlap2022using} introduces ALIA, a technique integrating both vision and language models. Using natural language descriptions of a dataset's classes or domains, ALIA edit the image using image-to-image diffusion models~\citep{brooks2023instructpix2pix}, ensuring the augmented data is not only visually consistent with the original but also encompasses a broader range of diversity, particularly beneficial for fine-grained classification tasks. \citep{trabucco2023effective} propose to change the inherent semantics of images, generalizing to novel visual concepts from a few labeled examples, making it especially valuable for tasks demanding semantic diversification.

\cm{Robust finetuning of vision models}, particularly the cutting-edge variants, has witnessed significant progress in recent times. Notably, image-text pre-trained models like CLIP~\citep{radford2021learning} have heralded unprecedented levels of robustness, as demonstrated in CLIP and subsequent studies \citep{wortsman2022robust,kumar2021fine}. While standard fine-tuning methodologies possess substantial potential, there's evidence to suggest that they might diminish robustness, especially in zero-shot paradigms. The culmination of methodologies from \citep{kumar2021fine} (LP-FT) and \citep{wortsman2022robust} (weight ensembling) represents a notable benchmark in the literature. Meanwhile, the approach by \citep{goyal2023finetune} introduces a nuanced strategy to the fine-tuning landscape. Harnessing a simple yet effective technique that mimics contrastive pretraining, it casts downstream class labels as text prompts and then optimizes the contrastive loss between image embeddings and these prompt embeddings, terming it "contrastive finetuning". This method has achieved remarkable results, outstripping benchmarks in multiple areas such as distribution shifts, transfer learning, and few-shot learning. Especially on the WILDS-iWILDCam, the FLYP approach championed by \citep{goyal2023finetune} has set new performance standards, surpassing both traditional finetuning and existing state-of-the-art approaches. The research solidifies contrastive finetuning as a premier, intuitive strategy for the supervised finetuning of image-text models like CLIP.

\cm{Supervised learning via contrastive loss} has taken center stage in recent research undertakings. The methodology advocates for the fine-tuning of zero-shot models in a fashion similar to their pre-training phase by capitalizing on contrastive loss. Various studies, such as~\citep{khosla2020supervised} have investigated this concept in a fully supervised setting without the support of a pre-trained model. In contrast,~\citep{gunel2020supervised} ventured into the realm of fine-tuning vast language models, while~\citep{zhang2021unleashing} concentrated on vision-only models. A salient distinction in the approach becomes evident when considering the addition of loss functions: while certain works have paired contrastive loss with cross-entropy, it has been observed that integrating cross-entropy with FLYP loss might negatively impact results. Direct comparisons between the two loss functions have showcased the superior accuracy credentials of contrastive loss over cross-entropy.

\cm{Generalization aspects and theoretical understanding of data augmentation} is a less explored area. Data augmentation plays a pivotal role in boosting performance, especially in tasks such as image and text classification. \citep{wu2020generalization} delves into the reasons behind the efficacy of various augmentations, specifically linear transformations, within the context of over-parametrized linear regression. The study reveals that certain transformations can either enhance estimation by expanding the span of the training data or act as regularization agents. Based on these insights, the authors present an augmentation strategy that tailors transformations to the model's uncertainty about the transformed data, validating its potency across image and text datasets. On the other hand, \citep{lin2022good} offers a fresh perspective on data augmentation (DA), challenging traditional beliefs. While classic augmentations, like translations in computer vision, are thought to create new data from the same distribution, this fails to explain the success of newer techniques that dramatically shift this distribution. The study introduces a theoretical framework that posits that DA imposes implicit spectral regularization, achieved through manipulating the eigenvalues of the data covariance matrix and boosting its entire spectrum via ridge regression. This framework provides a profound understanding of DA's varying impacts on generalization, serving as a foundational platform for innovative augmentation design.

\cm{Other augmentation methods that use soft labels and sample weights.} There is a recent paper, Soft-Augmentation~\citep{liu2023soft}, which also considers soft labels/targets and soft sample weights (i.e., loss reweighting). However, we believe there are huge methodological differences between the two methods in how they model the soft labels and weights. These methodological distinctions lead to significant differences in applicability. Below, we elaborate on the methodological and applicability differences between the two approaches and provide empirical comparisons to further highlight the novelty and improved performance of our method.

Our \SAFLEX employs a learnable, augmentation-method agnostic, and more automatic and principled approach for generating soft labels and sample weights. In Soft-Augmentation, the authors implement a specific approach to generating soft labels, namely through label smoothing. Label smoothing modifies the indicator value ``1'' (representing the ground-truth class label) with $p = 1 - \alpha(\phi)$, where the adaptive smoothing factor $\alpha(\phi)$ is determined by the degree/strength $\phi$ of the specific sampled augmentation applied to input $x_i$. Notably, the remaining probability mass $\alpha(\phi)$ is uniformly distributed across all other class labels. The formula of $\alpha(\cdot)$ requires human modeling with domain expertise. And since different upstream augmentation methods have different definitions of the strength factor $\phi$, remodeling of $\alpha(\cdot)$ for each new augmentation method is required. The discussion in Soft-Augmentation mainly focuses on crop augmentations on images, which impressively draws insights from human visual classification experiments. Our \SAFLEX, in contrast, differs in these key aspects: (a) Flexible Soft Labels: \SAFLEX employs a more flexible approach to modeling soft labels, moving beyond label smoothing's limitations. We believe that uniformly distributing the probability mass across all classes may not always be the most effective strategy. This limitation of Soft-Augmentation is also acknowledged in the paper. (b) Learned Soft Labels and Sample Weights: In \SAFLEX, both soft labels and sample weights are learned from a bilevel optimization problem, which is agnostic to the type and strength of the upstream augmentation method. (c) Bilevel Optimization Problem: \SAFLEX confronts the inherent challenge of soft augmentation by framing it as a bilevel optimization problem. This approach represents the first rigorous formulation of the problem, underscoring an important theoretical contribution. Additionally, we introduce novel and efficient algorithms specifically designed to tackle this bilevel optimization challenge.

Our \SAFLEX approach offers broader applicability compared to the Soft-Augmentation method. Unlike Soft-Augmentation, which requires an explicit augmentation strength parameter $\phi$, \SAFLEX seamlessly integrates with any upstream data augmentation mechanism, including diffusion models that lack the strength parameter $\phi$. This versatility enables \SAFLEX to effectively handle a wider range of data types, including medical and tabular data. \SAFLEX demonstrates its versatility by effectively handling a variety of tasks, including (standard) classification, fine-grained classification, out-of-distribution (OOD) generalization, and self-supervised learning. This broad applicability is evident in our comprehensive experiments. Conversely, Soft-Augmentation primarily focuses on image classification, with specific emphasis on model occlusion performance and calibration error, thus limiting its applicability to a narrower range of tasks.
\section{Experimental Setups and Implementation Details}
\label{apd:implement}

In this section, we provide more details about the experimental setups and implementation details.

The experiments are conducted on 4 NVIDIA Tesla V100 GPUs with 32GB memory each.

For the hyperparameter setting of \SAFLEX algorithm, we usually set the penalty coefficient $\beta=0$, and only set it $\beta=1$ for experiments on the tabular datasets.
We often keep the temperature $\tau=0.01$, and only set it to be $\tau=0.1$ on the CLIP finetuning experiment.
We do not conduct hyperparameter search for the hyperparameters of \SAFLEX algorithm, and we believe the performance can be further improved by hyperparameter search.

We then describe the infomation of datasets. The information about tabular datasets are listed below.

\begin{table}[htbp!]
    \adjustbox{max width=1.0\textwidth}{%
    {\renewcommand{\arraystretch}{1.25}%
    {\Huge
    \centering
    \begin{tabular}{lcccccccc}
    \toprule
    Dataset & Task & \# Features & \# Categorical & \# Continuous & Dataset Size & \# Positives & \# of Neg. & \% of Positives \\
    \midrule
    Appetency & Binary & 39 & 3 & 36 & 494,021 & 97,278 & 396,743 & 19.69 \\
    Arrhythmia & Binary & 226 & 0 & 226 & 452 & 66 & 386 & 14.60 \\
    Click & Binary & 12 & 7 & 5 & 39,948 & 6,728 & 33,220 & 16.84 \\
    Credit & Binary & 29 & 0 & 29 & 284,807 & 492 & 284,315 & 0.17 \\
    QSAR & Binary & 41 & 0 & 41 & 1,055 & 356 & 699 & 33.74 \\
    Shrutime & Binary & 11 & 3 & 8 & 10,000 & 2,037 & 7,963 & 20.37 \\
    Volkert & Multiclass (10) & 147 & 0 & 147 & 58,310 & --- & --- & --- \\
    \bottomrule
    \end{tabular}
    }}}
    \caption{Statistics of the tabular datasets.}
\end{table}

\begin{table}[htbp!]
    \begin{tabular}{ll}
    \toprule
    Dataset & Download Link \\
    \midrule
    Appetency & http://kdd.ics.uci.edu/databases/kddcup99 \\
    Arrhythmia & http://odds.cs.stonybrook.edu/arrhythmia-dataset/ \\
    Click & https://kdd.org/kdd-cup/view/kdd-cup-2012-track-2 \\
    Credit & https://www.kaggle.com/jacklizhi/creditcard \\
    QSAR & https://archive.ics.uci.edu/ml/datasets/QSAR+biodegradation \\
    Shrutime & https://www.kaggle.com/shrutimechlearn/churn-modelling \\
    Volkert & http://automl.chalearn.org/data \\
    \bottomrule
    \end{tabular}
    \caption{Links of the tabular datasets.}
\end{table}

The specific subsets of iWILDCam and CUB datasets used in diffusion-generated augmentation experiments are adopted form~\citep{dunlap2023diversify}.

Next, we show some more experiment results. The performance of Soft-Augmentation~\citep{liu2023soft} on MedMNIST datasets is listed below. Since the Soft-Augmentation paper focuses on improving crop augmentation and does not provide formulas to generate soft labels and sample weights for the upstream augmentations we considered, we test it with crop augmentation on the MedMNIST medical image datasets. We use the tuned hyperparameters for crop augmentation and Soft-Augmentation as described in the paper.

\begin{table}[htbp!]
\adjustbox{max width=1.0\textwidth}{%
{\renewcommand{\arraystretch}{1.30}%
{\Huge
\centering
\begin{tabular}{lcccccccc}
    \toprule
    \textbf{Method} & \textbf{Path} & \textbf{Derma} & \textbf{Tissue} & \textbf{Blood} & \textbf{OCT} & \textbf{OrganA} & \textbf{OrganC} & \textbf{OrganS} \\
    \midrule
    Crop & $92.68 \pm 0.82$ & $76.61 \pm 0.14$ & $67.38 \pm 0.19$ & $95.38 \pm 0.12$ & $77.50 \pm 0.11$ & $94.46 \pm 0.14$ & $90.29 \pm 0.09$ & $80.19 \pm 0.06$ \\
    Soft Augmentation (w/ Crop) & $91.95 \pm 0.59$ & $77.05 \pm 0.24$ & $67.06 \pm 0.44$ & $95.96 \pm 0.28$ & $76.92 \pm 0.46$ & $93.90 \pm 0.25$ & $91.44 \pm 0.24$ & $80.92 \pm 0.17$ \\
    \bottomrule
\end{tabular}
}}}
    \caption{Soft-Augment's performance on MedMNIST images, ResNet-18 backbone is used.}
    \label{tab:softaug}
\end{table}

We see that except on the DermaMNIST dataset, the performance of Soft-Augmentation is even lower than the No Augmentation baseline. While \SAFLEX's performance is consistently higher than the 'No Augmentation' baseline. Applying crop augmentation directly decreases the performance on most of the MedMNIST datasets. This is not surprising as we observed that applying RandAugment or Mixup directly also lowers the performance. However, the main reason for Soft-Augmentation's relatively poor performance is that it cannot consistently improve performance over the crop augmentation baseline (it shows improvement on Derma, Blood, OrganC, OrganS, but decreases performance on Path, Tissue, OCT, OrganA). This suggests that in situations with a high prevalence of poor-quality augmented samples (e.g., crop augmentation on medical images), Soft-Augmentation's relatively conservative strategy is inadequate in overcoming the significant noise and label errors introduced by these samples.

The performance of LP-A3~\citep{yang2022adversarial} on MedMNIST datasets (copied from the original paper) is listed below for reference.

\begin{table}[htbp!]
\adjustbox{max width=1.0\textwidth}{%
{\renewcommand{\arraystretch}{1.30}%
{\Huge
\centering
\begin{tabular}{lcccccccc}
    \toprule
    \textbf{Method} & \textbf{Path} & \textbf{Derma} & \textbf{Tissue} & \textbf{Blood} & \textbf{OCT} & \textbf{OrganA} & \textbf{OrganC} & \textbf{OrganS} \\
    \midrule
    LP-A3 & $94.42 \pm 0.24$ & $76.22 \pm 0.27$ & $68.63 \pm 0.14$ & $96.97 \pm 0.06$ & $80.27 \pm 0.54$ & $94.73 \pm 0.21$ & $92.41 \pm 0.22$ & $82.28 \pm 0.38$ \\
    \bottomrule
\end{tabular}
}}}
    \caption{LP-A3's performance on MedMNIST images, ResNet-18 backbone is used.}
    \label{tab:lpa3}
\end{table}

For the efficiency result, we found that on the eight MedMNIST datasets considered, the overhead of \SAFLEX measure as wall-clock time is 42\% on average, while more specifically, 54\% with RandAugment and 31\% with Mixup.
On the seven tabular datasets, on average the overhead of \SAFLEX is 81\% of the original training time per epoch.

\end{document}